\pdfoutput=1
\documentclass[11pt]{article}

\usepackage[]{naacl2021}

\usepackage{times}
\usepackage{latexsym}
\usepackage{amsmath, amssymb, amsfonts}
\usepackage{algorithm}
\usepackage{algorithmic}
\usepackage{float}
\usepackage{arydshln}
\usepackage{graphicx}
\usepackage[T1]{fontenc}
\usepackage[utf8]{inputenc}

\usepackage{microtype}

\newcommand{\R}{\mathbb{R}}

\title{Sequential Cross-Document Coreference Resolution}

\author{
    Emily Allaway \thanks{$^*$ Work done during internship at Amazon AI} $^1$ \And Shuai Wang $^2$\\
    $^1$ Department of Computer Science, Columbia University, New York, NY\\
    $^2$ Amazon AI\\
    \texttt{eallaway@cs.columbia.edu}\\
    \texttt{\{wshui, ballemig\}@amazon.com}\\
  \And Miguel Ballesteros $^2$ \\}

\begin{document}
\maketitle
\begin{abstract}
Relating entities and events in text is a key component of natural language understanding. Cross-document coreference resolution, in particular, is important for the growing interest in multi-document analysis tasks.
In this work we propose a new model that extends the efficient sequential prediction paradigm for coreference resolution to cross-document settings and 
achieves competitive results for both entity and event coreference while provides strong evidence of the efficacy of both sequential models and higher-order inference in cross-document settings.
Our model incrementally composes mentions into cluster representations and predicts links between a mention and the already constructed clusters,
approximating a higher-order model. 
In addition, we conduct extensive ablation studies that provide new insights into the importance of various inputs and representation types in coreference.
\end{abstract}

%%%%%%%%%%%%%%%%%%%
%% Intro
%%%%%%%%%%%%%%%%%%%
\section{Introduction}
Relating entities and events in text is a key component of natural language understanding. For example, 
whether two news articles describing hurricanes are referring to the same hurricane event. 
A crucial component of answering such questions is reasoning about and groups entities and events \textit{across multiple} documents.

The goal of coreference resolution is to compute these clusterings of entities or events from extracted spans of text. While within-document coreference has been studied extensively (e.g., ~\citet{Lee2017EndtoendNC,Lee2018HigherorderCR}), there has been relatively less work on the cross-document task. However, growing interest in multi-document applications, such as summarization (e.g.,~\citet{liu-lapata-2019-hierarchical,fabbri-etal-2019-multi}) and reading comprehension (e.g.,~\citet{Yan2019ADC,Welbl2018ConstructingDF}), highlights the importance of developing efficient and accurate cross-document coreference models to minimize error-propagation in complex reasoning tasks. 

In this work we focus on cross-document coreference (CDCR), which implicitly requires within-document coreference (WDCR), and propose a new model that improves both coreference performance and computational complexity. Recent advances in within-document entity coreference resolution have shown that sequential prediction (i.e., making coreference predictions from left to right in a text) achieves strong performance~\citep{Lee2017EndtoendNC} with lower computational costs. This paradigm is also well suited to real-world streaming settings,
where new documents are received every day, since it can easily find corefering events and entities with already processed documents, while most
non-sequential models
% graph-based approaches 
would require a full rerun. 
In this work, we show how this technique can first be extended to cross-document entity coreference and then adapted to cross-document event coreference.

Our method is also able to take advantage of the history of previously made coreference decisions, approximating a higher-order model (i.e., operating on mentions as well as structures with mentions). Specifically, for every mention, a coreference decision is made not over a set of individual mentions but rather over the \textit{current state of coreference clusters}. In this way, the model is able to use knowledge about the mentions currently in a cluster when making its decisions. 
While higher-order models 
have achieved state-of-the-art performance on entity coreference~\citep{Lee2018HigherorderCR}, 
they have been used infrequently for event coreference. For example,
~\citet{Yang2015AHD}
use one Chinese restaurant process for WDCR and then a second for CDCR over the within-document clusters. In contrast, our models make within- and cross-document coreference decisions in a single pass, taking into account all prior coreference decisions at each step.

Our contributions are: (1) we present the first study of sequential modeling for cross-document entity and event coreference and achieve competitive performance with a large reduction in computation time, (2) we conduct extensive ablation studies both on input information and model features, providing new insights for future models.

%%%%%%%%%%%%%%%%%%%
%% Related Work %%%
%%%%%%%%%%%%%%%%%%%
\section{Related Work}
Prior work on coreference resolution is generally split into either entity or event coreference. Entity coreference is relatively well studied~\cite{ng2010supervised}, with the largest focus on within-document coreference (e.g.,~\citet{raghunathan2010multi,fernandes2012latent,Durrett2013EasyVA,bjorkelund2014learning,martschat2015latent,wiseman2016learning,clark2016deep,Lee2018HigherorderCR,Kantor2019CoreferenceRW}). Recently,~\citet{Joshi2019BERTFC} showed that
pre-trained language models,
% BERT~\citep{Devlin2019BERTPO}, 
in particular BERT-large~\citep{Devlin2019BERTPO}, achieve state-of-the-art performance on entity coreference.
% and outperforms ELMo~\citep{Peters2018DeepCW}. 
In contrast to prior work on entity coreference, which is primarily sequential (i.e., left to right) and only within-document, our work extends the sequential paradigm to cross-document coreference and also adds incremental candidate composition.

There has been less work on event coreference since the task is generally considered harder. This is largely due to the more complex nature of event mentions (i.e., a trigger and arguments) and their syntactic diversity (e.g., both verb phrases and noun-phrases). Prior work on event coreference typically involves pairwise scoring between mentions followed by a standard clustering algorithm to predict coreference links~\citep{Pandian2018EventCR,Choubey2017EventCR,Cremisini2020NewII,Meged2020ParaphrasingVC,Yu2020PairedRL,Cattan2020StreamliningCC,Yu2020PairedRL}, classification over a fixed number of clusters~\citep{KenyonDean2018ResolvingEC} and template-based methods~\citep{Cybulska2015TranslatingGO,Cybulska2015BagOE}. While pairwise scoring (e.g., graph-based models, see \S \ref{sec:depparse}) with clustering are effective, they require tuned thresholds (for the clustering algorithm) and cannot use already predicted scores to inform later ones, since all scores are predicted independently. To the best of our knowledge, our work is the first to apply sequential models to cross-document event coreference.

Although a few previous works attempt to use information about existing clusters through incremental construction~\citep{Yang2015AHD,Lee2012JointEA} or argument sharing~\citep{Barhom2019RevisitingJM,Choubey2017EventCR}, these either continue to rely on pairwise decisions or use shallow, non-contextualized features that have limited efficacy. 
%EA: added
An exception is~\citet{Xu2020RevealingTM}, who explore a variant of cluster merging for WD entity conreference only and still relies on scores between individual mentions.
In contrast, our method makes coreference decisions between a mention and all \textit{existing coreference clusters} using contextualized features and so takes advantage of interdependencies between mentions while making all decisions in one pass.

%%%%%%%%%%%%%
%% Methods  %
%%%%%%%%%%%%%
\section{Methods}
\begin{figure*}
    \centering
    \vspace{-20pt}
    \includegraphics[width=0.85\textwidth]{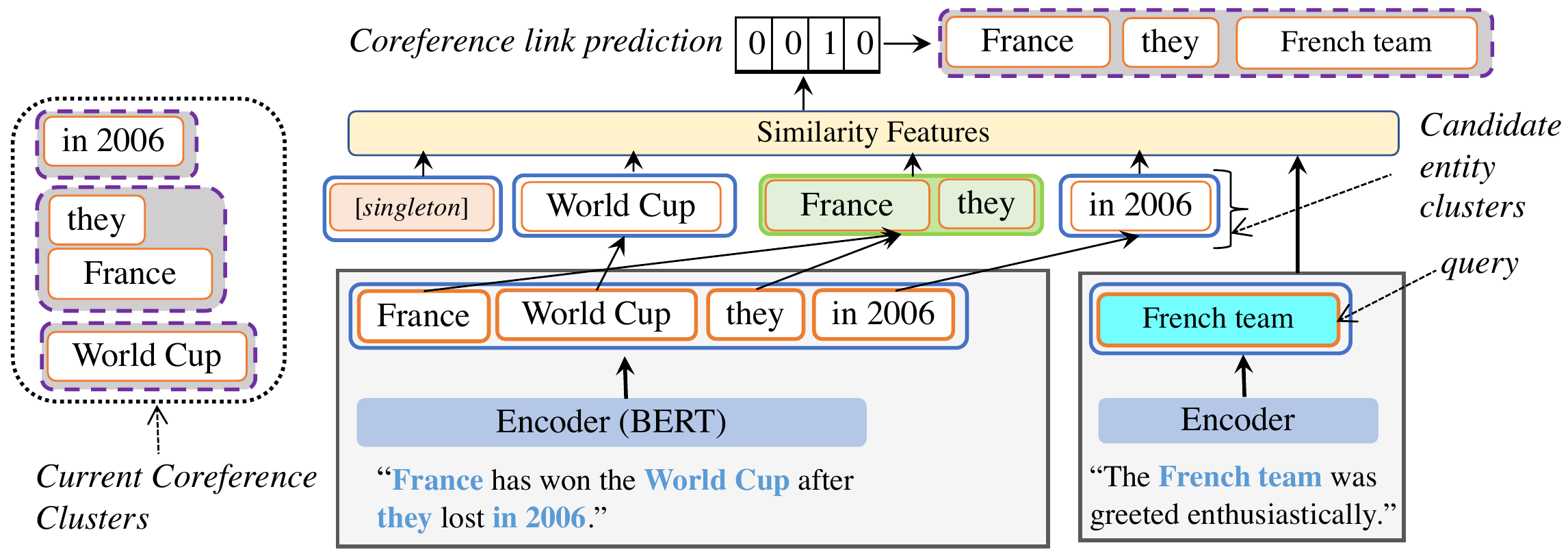}
    \caption{Our model using sequential prediction with incremental clustering for cross-document entity coreference.}
    \label{fig:model}
\end{figure*}
\subsection{Overview and Task Definition}
We propose a new sequential model for cross-document coreference resolution (see Figure~\ref{fig:model}) that predicts links between mentions and incrementally constructed coreference clusters computed across multiple documents. In the following sections, we will first describe our model for entity coreference (\S \ref{sec:entityrep}-\ref{sec:xdocpred}), then discuss adaptations to event coreference (\S \ref{sec:eventcoref}), and finally analyze the computational complexity of our algorithm and with an analogy to paradigms in dependency parsing (\S \ref{sec:depparse}).

The goal of entity coreference is to determine whether two entity mentions refer to the same real-world entity, with an analogous definition for event coreference. Formally, define an entity mention $x = \langle e, V\rangle$ where $e$ is an entity and $V$ is a set of events in which $e$ participates. We adopt the definition of an event as ``a specific occurrence of something that happens''~\citep{Cybulska2014UsingAS}. More specifically $V = [\langle t_1, r_1\rangle, \hdots, \langle t_n, r_n\rangle]$ where $t_i$ is an event trigger and $r_i \in R$ is the role $e$ takes in in the event with trigger $t_i$, from a fixed set of argument roles. 
 
\subsection{Entity Mention Representation}
\label{sec:entityrep}
To construct a representation for entity mention $x$, we first embed the entity $e$, along with its context, as
$h_e$ using the embeddings from BERT~\citep{Devlin2019BERTPO} of the start and end sub-word tokens of the entity span.
We similarly embed each event $v_i \in V$ as
$h_{v_i}$
Then we compute an aggregated event representation $h_v$ using a BiLSTM~\citep{Hochreiter1997LongSM} over all of the $h_{v_i}$, followed by mean-pooling.
Finally, we combine the entity representation and event representations using an affine transformation to obtain the full mention representation
$h_x$.

\subsection{Incremental Candidate Composition}
\label{sec:candidaterep}
Let $L^e = \{P_1, \hdots, P_n\}$ be a set of coreference clusters over the antecedents of mention $x_i$. 
We compute a candidate cluster representation $h_P$ for each set $P$ of coreferring entity antecedents in $L^e$.
In a similar manner to composition functions in neural dependency parsing, which incrementally combine head-word and modifier information to construct a subtree representation~\citep{Dyer2015TransitionBasedDP,Dyer2016RecurrentNN,Lhoneux2019RecursiveSC}, we incrementally combine document- and mention-level information to form a complete candidate cluster representation $h_P$. 
That is, for each $x_j \in P$,
we  combine $h_{x_j}$ and $h_{\text{CLS}_j}$, the CLS token embedding from the document containing $x_j$, using a non-linear transformation
\begin{equation}
    h_{c_j} = \tanh(W_x h_{x_j} + W_{\text{CLS}} h_{\text{CLS}_j} + b_c),
\label{eq:clscomp}
\end{equation}
where $W_x, W_{\text{CLS}} \in \R^{d_m \times d_m}$ and $b_c \in \R^{d_m}$ are learned parameters. 
Then we average the representations $h_{c_j}$ for all $x_j \in P$. 
To allow the model to predict singleton mentions, we add an additional candidate $S$, with representation $h_s = h_{\text{CLS}}$, to the set of candidates for $x_i$, where coreference with $S$ indicates $x_i$ is a singleton. 
As we update $L^e$, we incrementally update $h_P$ for all $P \in L^e$. Note that $L^e$ can be either the gold coreference clusters over seen mentions (during training) or the current set of predicted clusters (during inference). 

\subsection{Coreference Link Prediction}
\label{sec:linkpred}
We predict coreference links between a query entity mention $x$ and a set of candidates by passing a set of similarity features through a softmax layer.
Let $\mathcal{C}_x = \{h_{P_1}, \hdots, h_{P_m}\}$ be the set of $m$ candidate representations (including $h_s$) for $x$. 

We first compute the similarity between each candidate $P_j$ and the query using both cosine similarity $f_{cos}$ and multi-perspective cosine (MP cosine) similarity $f_{mpcos}$~\citep{Wang2017BilateralMM}. For multi-perspective cosine similarity, we first project the candidate and query into $k$ shared spaces using $k$ separate linear projections. Then, for each of the $k$ new spaces, we compute the cosine similarity between the projected representations in that space.

Next, we combine these features with the product and the difference of the candidate and query to obtain the final feature representation:
\begin{equation}
    h_f^{(j)} = [h_x \cdot h_{P_j}; |h_{x} - h_{P_j}|; f_{cos}; f_{mpcos}].
\label{eq:featrep}
\end{equation}

Then, for all candidates $P_j$ we compute the probability $p(x, P_j)$ that the query $x$ corefers with as
\begin{equation*}
    p(x \text{ corefers with } P_j) = softmax(W_o h_f + b_o).
\end{equation*}
We predict a link between $x$ and the candidate with maximum coreference probability. If that candidate is $S$, then $x$ is predicted as a singleton.
 
\subsection{Sequential Cross-Document Prediction}
\label{sec:xdocpred}
\begin{figure}[t]
    \input{tables/algorithm}
    \caption{Algorithm for coreference resolution with sequential prediction and incremental clustering (\S\ref{sec:candidaterep}).}
    \label{fig:algo}
\end{figure}
To predict cross-document coreference links, we propose an algorithm that iterates through a list of documents and predicts coreference links between entity mentions in the \textit{current document} and candidate clusters computed across all \textit{preceeding documents} (Figure \ref{fig:algo}). 

We first impose an arbitrary ordering on a list of documents $D$. Then, for each $i \in \{1, \hdots, |D|\}$ and each entity mention $x_n$ in document $D[i]$ we compute candidates clusters $\mathcal{C}_{x_n}$ (\S \ref{sec:candidaterep}) from the coreference clusters across all documents $D[j]$ where $j < i$. Note that this includes both within-document and cross-document clusters. 

After computing the candidate clusters for entity mention $x_n$, we predict a coreference link $\ell_n$ between $x_n$ and one candidate $P_j \in \mathcal{C}_{x_n}$ (\S \ref{sec:linkpred}). Finally, we update the predicted clustering to account for $\ell_n$ and re-compute new candidates for $x_{n + 1}$. 

Since the number of possible candidates for each $x_n$ grows as the number of preceding documents ($i$) increases, we reduce the computational cost by only considering previous documents $D[j]$ that are similar to $D[i]$. We define similar as having the same topic from a fixed set of topics $T$.

During training, we use gold entity clusters to compute the candidates (as in Figure~\ref{fig:algo}) and gold document topic clusters $T$. In contrast, during inference we using the currently predicted coreference clusters to compute candidates. That is, we use $L^e$ in place of $G^e$ in lines $9$ and $15$ in Figure~\ref{fig:algo}. Furthermore, we use predicted topic clusters $\Hat{T}$, computed using K-means (\S \ref{sec:impdetails}), in place of $T$. 

Our model is trained to minimize cross-entropy loss computed in batches. Here, all $M$ entity mentions in a single document form one batch and the loss is computed after $M$ sequential predictions.

\subsection{Adaptations for Event Coreference}
\label{sec:eventcoref}
We also adapt the same architecture and algorithm to cross-document event coreference resolution. Define an event $x = \langle t, A\rangle$ where $t$ is the event trigger and $A = [\langle e_1, r_1\rangle, \hdots, \langle e_m, r_m\rangle]$ is the set of its event arguments (i.e., entity-role pairs). If no entity takes some role $r_i$, then $e_i = \emptyset$.

We compute the event representation $h_x$ analogously to the entity representations (\S \ref{sec:entityrep}). That is, we combine the event trigger representation with an aggregated entity representation, computed over event arguments $A$. We then compute candidate-clusters and predict coreference links in the same manner as for entities (\S \ref{sec:candidaterep}, \S \ref{sec:linkpred}) with an additional feature, indicating whether event arguments corefer, in Equation~\ref{eq:featrep}.

Under the definition of event coreference, two events corefer when both their triggers and all of their arguments corefer. In practice, we relax the second requirement to \textit{most} of their arguments, since argument role labeling may be noisy. We compute a binary feature for $g_{r_l}$ for each argument role $r_l$ to indicate the coreference of $e_l$ (the entity with role $r_l$ in $x$) and $e_l^{(P_j)}$  (the entity with role $r_l$ in candidate cluster $P_j$). We compute a feature only for roles $r_l \in R$ in which both the candidate and the query have some entity present ($e_l^{(i)} \neq \emptyset$ and $e_l^{(P_j)} \neq \emptyset$). Then, for each $r_l \in R$, if the two entities corefer then $g_{r_l} = 1$ and if they do not corefer then $g_{r_l} = 0$. Finally, we map each $g_{r_l}$ to a learned embedding $f_{r_l} \in \R^{d_f}$ and compute an aggregated argument feature representation
$f_r = \frac{1}{|R_{\neq 0}|} \sum_{r_l \in R_{\neq 0}} f_{r_l}$
where $R_{\neq 0}$ is the set of roles filled in both $x$ and $P_j$. This feature is then concatenated into Equation~\ref{eq:featrep} before prediction. 

The cross-document iteration algorithm for event coreference is analogous to Figure~\ref{fig:algo} with the modification that ComputeFeatures (line 11) now also takes the gold entity coreference clusters $G^e$. 

\subsection{
Time Comparison with Prior Methods}
% Complexity Analysis and Analogy to Dependency Parsing}
\label{sec:depparse}
Algorithms for coreference resolution fall into two paradigms: sequential models (i.e., left to right prediction) and graph-based models (i.e., finding optimal connected components from a graph of pairwise similarity scores). This dichotomy is analogous to that in dependency parsing between transition-based parsers (i.e., greedy left-to-right models) and graph-based parsers. While the differences between the paradigms have been studied for dependency parsing~\citep{McDonald2007CharacterizingTE,McDonald2011AnalyzingAI}, comparisons for coreference have been limited to WD entity coreference only~\citep{martschat2015latent}. In part,  this is due to the usages of the two paradigms; sequential models are primarily used for WDCR while graph-based models are used for CDCR. However, as in dependency parsing, the sequential models can be made much more computationally efficient than the graph-based models as we show with our model. 

Let $D$ be a set of documents with $m$ mentions that form $c$ coreference clusters. In a \textit{graph-based} model, scores are computed between all pairs of mentions in all documents and in a \textit{general sequential} model scores are computed between a specific mention and all antecedents. Our model is a \textit{higher-order sequential} model; it combines the general sequential paradigm with higher-order inference through incremental candidate clustering. Therefore, while both general sequential and graph-based models \textit{always} require computing ${\sim}m^2$ scores, our model only needs to compute $cm$ scores. Since in practice $c\ll m$, our model is more efficient.

This improvement in efficiency is even more important in real-world streaming scenarios.
%In fact, our higher-order sequential model is also better suited to real-world streaming scenarios.
Namely, given a known set of clusters $C$ for the document set $D$, compute coreference with the mentions in a new document. In both a general sequential model and a graph-based model, scores need to be computed between the new mentions, and \textit{all} mentions in $D$. However, our model only needs to compare the new mentions to the existing \textit{clusters} $C$. Hence, our model can better handle the temporal-component of many usage settings and is better suited to life-long learning.

\section{Experimental Setup}
\subsection{Data}
\label{sec:data}
We conduct experiments using the ECB+ dataset~\cite{Cybulska2014UsingAS}, the largest available dataset for both within-document and cross-document event and entity coreference. The ECB+ dataset is an extension of the Event Coreference Bank dataset (ECB)~\citep{Bejan2010UnsupervisedEC}, which consists of news articles clustered into topics by seminal events (e.g., ``6.1 earthquake Indonesia 2009''). The extension of ECB adds an additional seminal event to each topic (e.g., ``6.1 earthquake Indonesia 2013''). Documents on each of the two seminal events then form subtopic clusters within each topic in ECB+. 

Following the recommendations of~\citet{Cybulska2015TranslatingGO}, we use only the subset of annotations that have been validated for correctness in our experiments (see Table~\ref{tab:datastats}). As a result, our results are comparable to recent studies (e.g., ~\citet{Barhom2019RevisitingJM,KenyonDean2018ResolvingEC,Meged2020ParaphrasingVC}) but not earlier methods (see~\citet{Upadhyay2016RevisitingTE} for a more complete overview of evaluation settings). We use the standard partitions of the dataset into train, development and test split by topic and use subtopics (gold or predicted) for document clustering.

\begin{table}[t]
    \centering
    \scalebox{0.9}{
    \begin{tabular}{lllll}
    \hline
        & Train & Dev & Test\\ \hline
        \# Topics & 25 & 8 & 10\\
        \# Subtopics & 50 & 16 & 20\\
        \# Documents & 574 & 196 & 206\\
        \# Event Mentions & 3808 & 1245 & 1780\\
        \# Entity Mentions & 4758 & 1476 & 2055\\
        \# Event Clusters & 1527 &  409 & 805\\
        \# Entity Clusters & 1286 & 330 & 608\\
    \hline
    \end{tabular}
    }
    \caption{Data Statistics for ECB+ corpus. Topics: train $\{1, 3, 4, 6$-$11, 13, 14, 16, 19$-$20, 22, 24$-$33\}$, development $\{2, 5, 12, 18, 21, 23, 34, 35\}$. and test $36$-$45$
    }
    \label{tab:datastats}
\end{table}

\subsection{Identifying Event Structures}
\label{sec:eventstruct}
The ECB+ dataset does not include relations between events and entities. Although prior work used the Swirl~\citep{Surdeanu2007CombinationSF} semantic role labeling (SRL) parser to extract predicate-argument structures, this does not take advantage of recent advances in SRL. In fact, prior works on coreference using ECB+ have added a number of additional rules on top of the parser output to improve its coverage and linking. For example,~\citet{Barhom2019RevisitingJM} used a dependency parser to identify additional mentions. Therefore, in this work we use the current state-of-the-art SRL parser on the standard CoNLL-2005 shared task~\citep{He2018JointlyPP}, which has improved performance by ${\sim}10$ F1 points both in- and out-of-domain. 

Following prior work, we restrict the event structure to the following four argument roles: ARG$0$, ARG$1$, TIME, and LOC. However, we additionally add a type constraint during pre-processing that requires entities of type TIME and LOC only fill matching roles (TIME and LOC respectively).

\subsection{Domain Adaptive Pre-training}
Since BERT was trained on the BooksCorpus and Wikipedia~\citep{Devlin2019BERTPO} and the ECB+ dataset contains news articles, there is a domain mismatch. In addition, the use of a domain corpus for pre-training helps address the data scarcity issue for events and entities, indicated by~\citet{Ma2020ResourceEnhancedNM}. Therefore, before training our coreference models, we first fine-tune BERT using the English Gigaword Corpus\footnote{https://catalog.ldc.upenn.edu/LDC2011T07} with both BERT losses, as this has been shown to be effective for domain transfer~\citep{Gururangan2020DontSP}. Following~\citet{Ma2020ResourceEnhancedNM}, we randomly sample $50k$  documents  ($626k$  sentences) and pre-train for $10k$ steps, using the hyperparameter settings from~\citet{Devlin2019BERTPO}. 

\subsection{Baselines and Models}
We experiment with the following baseline variations of our model: 
% \textbf{BERT-Rep} 
\textbf{BERT-SeqWD}
-- computes coreference scores using only the entity (or event) representations, without any cross-document linking, and 
% \textbf{BERT-Rep-Xdoc} 
\textbf{BERT-SeqXdoc}
-- computes coreference scores across documents but without candidate aggregation. 
% Note, both baselines 
% % We 
% are sequential and 
% use adaptive pre-training.
% % in both baselines. 
 For both event and entity coreference we experiment with our model, 
% \textbf{SeqXdoc}, 
\textbf{SeqXdoc+IC}
with (\textbf{+Adapt}) and without adaptive pre-training.

For entity coreference we compare against the following models:~\citet{Barhom2019RevisitingJM} (Bh2019) -- graph-based model that alternates between event and entity coreference, updating the argument features for events (and event features for entities) after each iteration, ~\citet{Cattan2020StreamliningCC} --  general sequential WDCR model from~\citet{Lee2017EndtoendNC} paired with agglomerative clustering, ~\citet{Caciularu2021CrossDocumentLM} -- graph-based Bh2019 model using representations from a Longformer~\citep{Beltagy2020LongformerTL} with cross-document attention during pre-training, and Lemma -- a strong baseline that links mentions with the same head-word lemma in the same document topic cluster~\citep{Barhom2019RevisitingJM}.

For event coreference we additionally compare to:
\citet{Meged2020ParaphrasingVC} -- graph-based Bh2019 model with an additional paraphrase-based feature, ~\citet{Cremisini2020NewII}
-- graph-based prediction of coreference scores over four types of similarity features, and three graph-based representation learning models with agglomerative clustering -- ~\citet{KenyonDean2018ResolvingEC},~\citet{Yu2020PairedRL} and~\citet{Zeng2020EventCR}.

% the prior state-of-the-art for entity coreference,~\citet{Cattan2020StreamliningCC}, and the current state-of-the-art for event coreference,~\citet{Yu2020PairedRL}, both of which use pairwise mention scoring followed by cross-document agglomerative clustering. 
% We note that both are contemporaneous with our work\footnote{Appearing on arXiv one month before our submission.}.

% In addition, we compare to three prior models from~\citet{Barhom2019RevisitingJM} for both entity and event coreference and three additional models for event coreference only. Specifically, for both entity and event  coreference we compare to: Joint -- alternates between event and entity coreference, updating the argument features for events (and event features for entities) after each iteration, Disjoint -- the joint model without argument features for event coreference (or event features for entity coreference), and Lemma -- a strong baseline that links mentions with the same head-word lemma in the same document topic cluster~\citep{Barhom2019RevisitingJM}. For event coreference only, we compare to:~\citet{Meged2020ParaphrasingVC} -- uses Joint model with an additional paraphrase-based feature, ~\citet{Cremisini2020NewII}
% -- predicts coreference scores over four types of similarity features without argument information, and~\citet{KenyonDean2018ResolvingEC} -- trains event representations without argument information.

\subsection{Implementation Details}
\label{sec:impdetails}
Our models are tuned for a maximum of $80$ epochs with early-stopping on the development set (using CoNLL F1) with a patience of $20$ epochs. All models are optimized using Adam~\citep{Kingma2015AdamAM} with a learning rate of $2e$-$5$ and treat all mentions in a document as a batch.
We apply gradient clipping to $30$ prevent exploding gradients. We encode each document using BERT-base and a maximum document length of $600$ tokens for BERT. Following adaptive pre-training, we do not fine-tune BERT. To encode arguments/events we use an LSTM with hidden size $128$, for the argument coreference features we use two learned embeddings of dimension $d_f = 50$, and for the multi-perspective cosine similarity we use $k=1$ projection layers with dimension $50$ for entity coreference and $k=3$ projection layers with the same dimension for event coreference.
% \footnote{We train all models with one Tesla V100-SXM2 GPU.}

We follow~\citet{Barhom2019RevisitingJM} and use K-means to compute document clusters for inference from their implementation with $K=20$. Specifically, as features, we use the TF-IDF scores of unigrams, bigrams, and trigrams in the unfiltered dataset, excluding stop words.

%%%%%%%%%%%%%%%%%%
%% Results
%%%%%%%%%%%%%%%%%%
\section{Results and Analyses}
\subsection{Coreference Resolution}
We evaluate using the standard metrics for coreference resolution: MUC~\citep{Vilain1995AMC}, $B^3$~\citep{Bagga1998AlgorithmsFS}, CEAF-$e$~\citep{Luo2005OnCR}, and CoNLL F1 (their average).

\begin{table*}[t]
    \centering
    \scalebox{0.95}{
    \begin{tabular}{l|rrr|rrr|rrr|r}
        \hline
        & \multicolumn{3}{c|}{MUC} & \multicolumn{3}{c|}{$B^3$} & \multicolumn{3}{c|}{CEAF-$e$} & \\ 
        & P & R & F1 & P & R & F1 & P & R & F1 & C-F1\\ \hline
        %  \hline
        % \citet{Barhom2019RevisitingJM}-
        Lemma
        % Bh2019-Lemma 
        & 71.3 & 83 & 76.7 & 53.4 & 84.9 & 65.6 & 70.1 & 52.5 & 60.0 & 67.4\\
        % Bh2019-Disjoint 
        % \citet{Barhom2019RevisitingJM}-Disjoint
        % & 76.7 & 80.8 & 78.7 & 63.2 & 78.2 & 69.9 & 65.3 & 58.3 & 61.6 & 70.0\\
        % Bh2019-Joint 
        \citet{Barhom2019RevisitingJM}
        % -Joint
        & 78.6 & 80.9 & 79.7 & 65.5 & 76.4 & 70.5 & 65.4 & 61.3 & 63.3 & 71.2\\ 
        % \hdashline
        \citet{Cattan2020StreamliningCC} & 85.7 & 81.7 & 83.6 & 70.7 & 74.8 & 72.7 & 59.3 & 67.4 & 63.1 & 73.1 \\ 
        \citet{Caciularu2021CrossDocumentLM} & \textbf{88.1} & \textbf{91.8} & \textbf{89.9} & \textbf{82.5} & \textbf{81.7} &\textbf{ 82.1} & \textbf{81.2} & \textbf{72.9} & \textbf{76.8} & \textbf{82.9} \\
        \hline
        % BERT-Rep 
        BERT-SeqWD + Adapt
        & 78.0 & 39.2 & 52.2 & 89.6 & 34.5 & 49.8 & 34.9 & 76.1 & 47.9 & 50.0\\
        % BERT-Rep-Xdoc 
        BERT-SeqXdoc + Adapt
        & 80.2 & 69.8 & 74.6 & 76.6 & 54.2 & 63.5 & 49.6 & 64.8 & 56.2 & 64.8\\
        \hdashline
        SeqXdoc+IC
        & 83.6 & 81.5 & 82.5 & \underline{76.0} & 66.7 & 71.1 & 65.7 & \underline{69.3} & 67.4 & 73.7\\ 
        + Adapt &  \underline{83.9} & \underline{84.7} & \underline{84.3} & 74.5 & \underline{70.5} & \underline{72.4} & \underline{70.0} & 68.1 & \underline{69.2} & \underline{75.3} \\
        \hline
    \end{tabular}
    }
    \caption{Entity coreference on the ECB+ test set, combined within- and cross-document scores using predicted document clusters. C-F1 is CoNLL F1. \textbf{Bold} indicates best overall, \underline{underline} indicates our best model.}
    \label{tab:entitycoref}
\end{table*}
\begin{table*}[t]
    \centering
    \scalebox{0.95}{
    \begin{tabular}{l|rrr|rrr|rrr|r}
        \hline
        & \multicolumn{3}{c|}{MUC} & \multicolumn{3}{c|}{$B^3$} & \multicolumn{3}{c|}{CEAF-$e$} & \\ 
        & P & R & F1 & P & R & F1 & P & R & F1 & C- F1\\ \hline
        %  \hline
        % Bh2019-Lemma 
        % \citet{Barhom2019RevisitingJM}-
        Lemma
        & 76.5 & 79.9 & 78.1 & 71.7 & 85.0 & 77.8 & 75.5 & 71.7 & 73.6 & 76.5 \\
        % Kd2018 
        \citet{KenyonDean2018ResolvingEC}
        & 67.0 & 71.0 & 69.0 & 71.0 & 67.0 & 69.0 & 71.0 & 67.0 & 69.0 & 71.0 \\
        % Bh2019-Disjoint
        % \citet{Barhom2019RevisitingJM}-Disjoint
        % & 75.5 & 83.6 & 79.4 & 75.4 & 86.0 & 84.4 & 80.3 & 71.9 & 75.5 & 78.5 \\
        % Bh2019-Joint
        \citet{Barhom2019RevisitingJM}
        % -Joint
        & 77.6 & 84.5 & 80.9 & 76.1 & 85.1 & 80.3 & 81.0 & 73.8 & 77.3 & 79.5 \\
        % Cr2020 
        \citet{Cremisini2020NewII}
        & \textbf{89.4} & 84.9 & 87.1 & 74.3 & 69.2 & 71.6 & 49.6 & 60.7 & 54.6 & 71.1 \\
        % Mg2020 
        \citet{Meged2020ParaphrasingVC}
        & 78.7 & 84.7 & 81.6 & 75.9 & 85.9 & 80.5 & 81.1 & 74.8 & 77.8 & 80.0 \\ 
        % \hdashline
        % Ca2020 
        \citet{Cattan2020StreamliningCC}
        & 85.1 & 81.9 & 83.5 & 82.1 & 82.7 & 82.4 & 75.2 & 78.9 & 77.0 & 81.0 \\
        % Yu2020 
        \citet{Yu2020PairedRL}
        & 88.1 & 85.1 & 86.6 & \textbf{86.1} & 84.7 & 85.4 & 79.6 & \textbf{83.1} & 81.3 &  84.4 \\
          \citet{Zeng2020EventCR} & 85.6 & \textbf{89.3} & 87.5 & 77.6 & \textbf{89.7} & 83.2 & \textbf{84.5} & 80.1 & \textbf{82.3} & 84.3\\
          
          \citet{Caciularu2021CrossDocumentLM} & 87.1 & 89.2 & \textbf{88.1} & 84.9 & 87.9 & \textbf{86.4} & 83.3 & 81.2 & 82.2 & \textbf{85.6} \\
        
        \hline
    %   BERT-Rep 
       BERT-SeqWD + Adapt
       & 68.9 & 28.9 & 40.7 & 91.1 & 48.5 & 63.3 & 49.3 & 83.9 & 62.1 & 55.4\\
        % BERT-Rep-Xdoc 
        BERT-SeqXdoc + Adapt
        & 82.2 & 66.8 & 73.7 & 84.2 & 66.8 & 74.5 & 65.9 & 80.8 & 72.6 & 73.6\\
        \hdashline
        SeqXdoc+IC 
        & 81.6 & \underline{85.9} & \underline{83.7} & 69.5 & 80.6 & 74.4 & 75.3 & 67.1 & 71.0 & 76.4
        \\ 
        + Adapt &  \underline{81.7} & 82.8 & 82.2 & \underline{80.8} & \underline{81.5} & \underline{81.1} & \underline{79.8} & \underline{78.4} & \underline{79.1} & \underline{80.8}\\
        \hline
    \end{tabular}}
    \caption{Event coreference on the ECB+ test set, combined within- and cross-document scores using predicted document clusters. C-F1: is CoNLL F1. \textbf{Bold} indicates best overall, \underline{underline} indicates our best model. }
    \label{tab:eventcoref}
\end{table*}

For both entity coreference, our model outperforms most prior methods (see Table~\ref{tab:entitycoref}) and for event coreference our model demonstrates strong performance (see Table~\ref{tab:eventcoref}). Although~\citet{Caciularu2021CrossDocumentLM} achieve the highest performance, we observe that most of their gains come from the use of a Longformer (80.4 CoNLL F1 for entities, 84.6 for events) -- a language-model specifically designed for long contexts. As observed by~\citet{Xu2020RevealingTM}, improvements in the underlying language model can result in large-gains in coreference performance without requiring algorithmic improvements. However, our model focuses on improving the coreference prediction algorithm, while using a standard BERT language-model. 

We observe that our algorithm provides large gains for both event and entity coreference. In particular, while naive applications of the sequential paradigm in the cross-document setting (BERT-SeqWD and BERT-SeqXdoc) performs poorly, the addition of incremental candidate clustering even without adaptive pre-training yields competitive results ($+8.9$ and $+2.8$ CoNLL F1 for entities and events respectively). Adaptive pre-training, which handles domain-mismatch in a similar way to task-specific fine-tuning (e.g., as in~\citet{Caciularu2021CrossDocumentLM}), provides further gains ($4.1$ and $2.2$ CoNLL F1 for entities and events respectively). Therefore, our results highlight the 
importance of higher-order inference (e.g., composition) when extending 
sequential prediction to cross-document settings.

We note that prior work~\citep{Barhom2019RevisitingJM} used predicted entity coreference clusters. In a comparable setting, using the output from our best entity coreference model to compute argument coreference features (\S \ref{sec:linkpred}), we do not observe any drop in performance (i.e., the performance is identical). In addition, we use predicted document clusters for our experiments on both entity and event coreference. Due to the high-quality document clustering (Homogeneity: 0.977, Completeness: .980, V-measure: .978, Adjusted Random-Index: .945), we only observe a drop of ${\sim}1$ CoNLL F1 point when using these predicted clusters, compared to the gold document clusters. 
% Specifically, when using gold document clusters, our model obtains $76.0$ and $82.2$ CoNLL F1 for entity and event coreference respectively. 
However, we note that such a small decrease relies on the quality of the clustering, as shown by the larger gap ($3$ F1 points) observed by~\citet{Cremisini2020NewII} with less accurate clusters. 

Finally, we observe that the sequential paradigm with incremental candidate composition has several additional advantages. First, without candidate composition, sequential coreference resolution is typically a multi-label task. However, \textit{with} composition, each mention now has exactly one correct coreferring antecendent during training (possibly a cluster rather than an individual mention) and this simplifies the learning task. While heuristics (e.g., choosing the closest antecedent as the gold antecendent, discussed in~\citet{martschat2015latent}), also convert the task from multi- to single-label, they are problematic because they limit the amount of information that can be used during training. Additionally, candidate composition allows sequential models to make use of information not only about antecedents (in contrast to the graph-based models) but also about prior coreference decisions (in contrast to non-compositional sequential models). 

Our results are consistent with prior work on the efficacy of sequential models (cf. mention-ranking models for WD entity coreference)~\citep{martschat2015latent} and the importance of higher-order inference mechanisms (e.g., incremental candidate clustering) in cross-document tasks~\citep{Zhou2020MultilevelTA}. In addition, our results demonstrate the importance of algorithmic improvements, in addition to improvements in the underlying language model, for strong coreference performance.

%%%%%%%%%%%%%%%%%%
%% Analysis
%%%%%%%%%%%%%%%%%%
% \section{Analysis}
\subsection{Feature Ablation}
\label{sec:featablation}
\begin{table}[t]
    \centering
    \scalebox{0.97}{
    \begin{tabular}{l|rr|rr}
        \hline
         & \multicolumn{2}{c|}{Entity} & \multicolumn{2}{c}{Event}\\
         & F1 & $\Delta$ & F1 & $\Delta$ \\ \hline
        Our Model &  75.3 & & 80.8 &\\ \hdashline
        $-$ Coref feat (\S\ref{sec:eventcoref}) & - & - & 79.6 & -1.2\\
        $-$ Args (\S\ref{sec:entityrep}) & 74.8 & -0.9 & 78.7 & -2.1\\
        $-$ Arg comp (\S\ref{sec:entityrep}) & 74.6 & -0.7 & 78.3 & -2.5 \\
        $-$ CLS (Eq.~\ref{eq:clscomp}) & 74.5 & -0.8 & 78.9 & -1.9\\
        $-$ MP cosine (\S\ref{sec:linkpred}) & 74.5 & -0.8 & 79.1 & -1.7 \\ \hdashline
        $+$ GloVE & 70.1 & -5.2 & 76.7 & -4.1\\ 
        $+$ RoBERTa & 71.2 & -4.1 & 78.1 & -2.7\\ 
        \hline
    \end{tabular}
    }
    \caption{Ablation results (CoNLL F1) for both used and unused features on the ECB+ test set. For entity coreference arguments (Args) are events, while for event coreference they are entities.}
    \label{tab:featablation}
\end{table}
Since mention representations in coreference vary widely, we conduct extensive feature ablations to provide insights for future work (see Table~\ref{tab:featablation}).

First we examine the vector representations used to encode mentions. While prior work used ELMo and pre-trained GloVE~\citep{Pennington2014GloveGV} word and character embeddings, recent models use RoBERTa~\citep{Cattan2020StreamliningCC,Yu2020PairedRL}. We experiment with replacing BERT-base with RoBERTa-base and with using GloVE in addition BERT in our models (see Appendix~\ref{sec:appendix} for implementation) and observe large drops in performance. We hypothesize that without task-specific fine-tuning, adaptive pre-training is more beneficial for coreference on ECB+. 

We also observe that our entity coreference model is relatively less susceptible to feature changes than the event coreference model. For example, the event coreference model is particularly reliant on the argument features. Specifically, replacing the argument composition BiLSTM with a single mean-pooling operation ($-$ Arg comp) and removing all argument information ($-$ Args) both result in large drops in performance ($-2.5$ and $-2.1$ respectively). 

Finally, the contribution of the multi-perspective cosine similarity underscores the importance of cosine similarity as observed by~\citet{Cremisini2020NewII}. These ablations, including on the importance of document-level information ($-$ CLS) suggest new directions for token and document representations in coreference. 

\subsection{Effects of SRL}
\begin{table}[t]
    \centering
    \begin{tabular}{l|rr|rr}
        \hline
         & \multicolumn{2}{c|}{Entity} & \multicolumn{2}{c}{Event}\\
         & F1 & $\Delta$ & F1 & $\Delta$ \\ \hline
         Our Model & 75.3 &  & 80.8 & \\ \hdashline
         HeSRL - C & 75.3 & -0.0 & 80.4 & -0.4 \\
         HeSRL + BhR + C & 74.9 & -0.4 & 79.2 & -1.6 \\ \hdashline
         Swirl + BhR + C & 75.4 & +0.1 & 80.0 & -0.8\\
         Swirl + BhR & 75.4 & +0.1 & 78.7 & -2.1\\
         \hline
    \end{tabular}
    \caption{Ablation results  (CoNLL F1) on methods for identifying event structures on ECB+ test set. HeSRL is~\citet{He2018JointlyPP}, BhR is additional rules for aligning the SRL and annotations from~\citep{Barhom2019RevisitingJM}, C is entity type constraint (see \S \ref{sec:eventstruct}).}
    \label{tab:srlablation}
\end{table}
We investigate the impact of using a recent SRL parser to extract event structures (\S \ref{sec:eventstruct}), compared to the Swirl parser used in prior work (see Table~\ref{tab:srlablation}). 

We first observe that the additional extraction rules used in~\citet{Barhom2019RevisitingJM} are not necessary when using the new SRL parser. In fact, these rules actually result in a decrease in performance for both entity and event coreference ($-1.6$ and $-0.4$ respectively).
In addition, when using the Swirl parser and additional rules (Swirl+Bh-rules), we observe a large drop for event coreference ($-2.1$) compared to entity coreference. This aligns with the heavier dependence of event coreference models on arguments (\S~\ref{sec:featablation}), which will lead to greater model sensitivity to errors in the entity-event structures (from the SRL). 
Furthermore, we also see that the type constraint improves event coreference more when using the Swirl SRL ($\Delta = 1.3$) than when using the new SRL ($\Delta = 0.4$). Note that because we do not use role information for entity coreference (i.e., no argument coreference feature), adding or removing the type constraint does not affect entity coreference. These results highlight the importance of minimizing error propagation from the SRL into the coreference resolution.

%%%%%%%%%%%%%%%%%%
%% Conclusion
%%%%%%%%%%%%%%%%%%
\section{Conclusion}
In this paper, we propose a new model for cross-document coreference resolution that extends the efficient sequential prediction paradigm to multiple documents. The sequential prediction is combined with incremental candidate composition that allows the model to use the history of past coreference decisions at every step. 
Our model achieves 
competitive results for both entity and event coreference and provides strong evidence of the efficacy of both sequential models and higher-order inference in cross-document settings.
% state-of-the-art performance on cross-document entity coreference and competitive results for event coreference. 
In addition, we conduct extensive ablations and provide insights into better document- and mention-level representations. In future, we intend to adapt this model to coreference across document streams and investigate alternatives to greedy prediction (e.g., beam search).

% \section*{Acknowledgments}

% Entries for the entire Anthology, followed by custom entries
\bibliography{anthology,custom}

\begin{thebibliography}{54}
\expandafter\ifx\csname natexlab\endcsname\relax\def\natexlab#1{#1}\fi

\bibitem[{Bagga and Baldwin(1998)}]{Bagga1998AlgorithmsFS}
A.~Bagga and B.~Baldwin. 1998.
\newblock Algorithms for scoring coreference chains.

\bibitem[{Barhom et~al.(2019)Barhom, Shwartz, Eirew, Bugert, Reimers, and
  Dagan}]{Barhom2019RevisitingJM}
Shany Barhom, Vered Shwartz, Alon Eirew, Michael Bugert, Nils Reimers, and
  I.~Dagan. 2019.
\newblock Revisiting joint modeling of cross-document entity and event
  coreference resolution.
\newblock In \emph{ACL}.

\bibitem[{Bejan and Harabagiu(2010)}]{Bejan2010UnsupervisedEC}
C.~Bejan and Sanda~M. Harabagiu. 2010.
\newblock Unsupervised event coreference resolution with rich linguistic
  features.
\newblock In \emph{ACL}.

\bibitem[{Beltagy et~al.(2020)Beltagy, Peters, and
  Cohan}]{Beltagy2020LongformerTL}
Iz~Beltagy, Matthew~E. Peters, and Arman Cohan. 2020.
\newblock Longformer: The long-document transformer.
\newblock \emph{ArXiv}, abs/2004.05150.

\bibitem[{Bj{\"o}rkelund and Kuhn(2014)}]{bjorkelund2014learning}
Anders Bj{\"o}rkelund and Jonas Kuhn. 2014.
\newblock Learning structured perceptrons for coreference resolution with
  latent antecedents and non-local features.
\newblock In \emph{Proceedings of the 52nd Annual Meeting of the Association
  for Computational Linguistics (Volume 1: Long Papers)}, pages 47--57.

\bibitem[{Caciularu et~al.(2021)Caciularu, Cohan, Beltagy, Peters, Cattan, and
  Dagan}]{Caciularu2021CrossDocumentLM}
Avi Caciularu, Arman Cohan, Iz~Beltagy, Matthew~E. Peters, Arie Cattan, and Ido
  Dagan. 2021.
\newblock Cross-document language modeling.
\newblock \emph{ArXiv}, abs/2101.00406.

\bibitem[{Cai and Strube(2010)}]{Cai2010EvaluationMF}
Jie Cai and Michael Strube. 2010.
\newblock Evaluation metrics for end-to-end coreference resolution systems.
\newblock In \emph{SIGDIAL Conference}.

\bibitem[{Cattan et~al.(2020)Cattan, Eirew, Stanovsky, Joshi, and
  Dagan}]{Cattan2020StreamliningCC}
Arie Cattan, Alon Eirew, Gabriel Stanovsky, Mandar Joshi, and I.~Dagan. 2020.
\newblock Streamlining cross-document coreference resolution: Evaluation and
  modeling.
\newblock \emph{ArXiv}, abs/2009.11032.

\bibitem[{Choubey and Huang(2017)}]{Choubey2017EventCR}
Prafulla~Kumar Choubey and Ruihong Huang. 2017.
\newblock Event coreference resolution by iteratively unfolding
  inter-dependencies among events.
\newblock In \emph{EMNLP}.

\bibitem[{Clark and Manning(2016)}]{clark2016deep}
Kevin Clark and Christopher~D Manning. 2016.
\newblock Deep reinforcement learning for mention-ranking coreference models.
\newblock In \emph{Proceedings of the 2016 Conference on Empirical Methods in
  Natural Language Processing}, pages 2256--2262.

\bibitem[{Cremisini and Finlayson(2020)}]{Cremisini2020NewII}
Andres Cremisini and Mark~A. Finlayson. 2020.
\newblock New insights into cross-document event coreference: Systematic
  comparison and a simplified approach.
\newblock In \emph{NUSE}.

\bibitem[{Cybulska and Vossen(2014)}]{Cybulska2014UsingAS}
A.~Cybulska and Piek T. J.~M. Vossen. 2014.
\newblock Using a sledgehammer to crack a nut? lexical diversity and event
  coreference resolution.
\newblock In \emph{LREC}.

\bibitem[{Cybulska and Vossen(2015{\natexlab{a}})}]{Cybulska2015BagOE}
A.~Cybulska and Piek T. J.~M. Vossen. 2015{\natexlab{a}}.
\newblock "bag of events" approach to event coreference resolution. supervised
  classification of event templates.
\newblock \emph{Int. J. Comput. Linguistics Appl.}, 6:11--27.

\bibitem[{Cybulska and Vossen(2015{\natexlab{b}})}]{Cybulska2015TranslatingGO}
A.~Cybulska and Piek T. J.~M. Vossen. 2015{\natexlab{b}}.
\newblock Translating granularity of event slots into features for event
  coreference resolution.
\newblock In \emph{EVENTS@HLP-NAACL}.

\bibitem[{de~Lhoneux et~al.(2019)de~Lhoneux, Ballesteros, and
  Nivre}]{Lhoneux2019RecursiveSC}
Miryam de~Lhoneux, Miguel Ballesteros, and Joakim Nivre. 2019.
\newblock Recursive subtree composition in lstm-based dependency parsing.
\newblock \emph{ArXiv}, abs/1902.09781.

\bibitem[{Devlin et~al.(2019)Devlin, Chang, Lee, and
  Toutanova}]{Devlin2019BERTPO}
J.~Devlin, Ming-Wei Chang, Kenton Lee, and Kristina Toutanova. 2019.
\newblock Bert: Pre-training of deep bidirectional transformers for language
  understanding.
\newblock In \emph{NAACL-HLT}.

\bibitem[{Durrett and Klein(2013)}]{Durrett2013EasyVA}
Greg Durrett and D.~Klein. 2013.
\newblock Easy victories and uphill battles in coreference resolution.
\newblock In \emph{EMNLP}.

\bibitem[{Dyer et~al.(2015)Dyer, Ballesteros, Ling, Matthews, and
  Smith}]{Dyer2015TransitionBasedDP}
Chris Dyer, Miguel Ballesteros, W.~Ling, A.~Matthews, and Noah~A. Smith. 2015.
\newblock Transition-based dependency parsing with stack long short-term
  memory.
\newblock In \emph{ACL}.

\bibitem[{Dyer et~al.(2016)Dyer, Kuncoro, Ballesteros, and
  Smith}]{Dyer2016RecurrentNN}
Chris Dyer, Adhiguna Kuncoro, Miguel Ballesteros, and Noah~A. Smith. 2016.
\newblock Recurrent neural network grammars.
\newblock In \emph{HLT-NAACL}.

\bibitem[{Fabbri et~al.(2019)Fabbri, Li, She, Li, and
  Radev}]{fabbri-etal-2019-multi}
Alexander Fabbri, Irene Li, Tianwei She, Suyi Li, and Dragomir Radev. 2019.
\newblock \href {https://doi.org/10.18653/v1/P19-1102} {Multi-news: A
  large-scale multi-document summarization dataset and abstractive hierarchical
  model}.
\newblock In \emph{Proceedings of the 57th Annual Meeting of the Association
  for Computational Linguistics}, pages 1074--1084, Florence, Italy.
  Association for Computational Linguistics.

\bibitem[{Fernandes et~al.(2012)Fernandes, dos Santos, and
  Milidi{\'u}}]{fernandes2012latent}
Eraldo Fernandes, Cicero dos Santos, and Ruy~Luiz Milidi{\'u}. 2012.
\newblock Latent structure perceptron with feature induction for unrestricted
  coreference resolution.
\newblock In \emph{Joint Conference on EMNLP and CoNLL-Shared Task}, pages
  41--48.

\bibitem[{Gururangan et~al.(2020)Gururangan, Marasovi{\'c}, Swayamdipta, Lo,
  Beltagy, Downey, and Smith}]{Gururangan2020DontSP}
Suchin Gururangan, Ana Marasovi{\'c}, Swabha Swayamdipta, Kyle Lo, Iz~Beltagy,
  Doug Downey, and Noah~A. Smith. 2020.
\newblock Don't stop pretraining: Adapt language models to domains and tasks.
\newblock In \emph{ACL}.

\bibitem[{He et~al.(2018)He, Lee, Levy, and Zettlemoyer}]{He2018JointlyPP}
Luheng He, Kenton Lee, Omer Levy, and Luke Zettlemoyer. 2018.
\newblock Jointly predicting predicates and arguments in neural semantic role
  labeling.
\newblock In \emph{ACL}.

\bibitem[{Hochreiter and Schmidhuber(1997)}]{Hochreiter1997LongSM}
S.~Hochreiter and J.~Schmidhuber. 1997.
\newblock Long short-term memory.
\newblock \emph{Neural Computation}, 9:1735--1780.

\bibitem[{Joshi et~al.(2019)Joshi, Levy, Weld, and
  Zettlemoyer}]{Joshi2019BERTFC}
Mandar Joshi, Omer Levy, Daniel~S. Weld, and Luke Zettlemoyer. 2019.
\newblock Bert for coreference resolution: Baselines and analysis.
\newblock In \emph{EMNLP/IJCNLP}.

\bibitem[{Kantor and Globerson(2019)}]{Kantor2019CoreferenceRW}
Ben Kantor and A.~Globerson. 2019.
\newblock Coreference resolution with entity equalization.
\newblock In \emph{ACL}.

\bibitem[{Kenyon-Dean et~al.(2018)Kenyon-Dean, Cheung, and
  Precup}]{KenyonDean2018ResolvingEC}
Kian Kenyon-Dean, J.~Cheung, and Doina Precup. 2018.
\newblock Resolving event coreference with supervised representation learning
  and clustering-oriented regularization.
\newblock \emph{ArXiv}, abs/1805.10985.

\bibitem[{Kingma and Ba(2015)}]{Kingma2015AdamAM}
Diederik~P. Kingma and Jimmy Ba. 2015.
\newblock Adam: A method for stochastic optimization.
\newblock \emph{CoRR}, abs/1412.6980.

\bibitem[{Lee et~al.(2012)Lee, Recasens, Chang, Surdeanu, and
  Jurafsky}]{Lee2012JointEA}
H.~Lee, M.~Recasens, Angel~X. Chang, M.~Surdeanu, and Dan Jurafsky. 2012.
\newblock Joint entity and event coreference resolution across documents.
\newblock In \emph{EMNLP-CoNLL}.

\bibitem[{Lee et~al.(2017)Lee, He, Lewis, and Zettlemoyer}]{Lee2017EndtoendNC}
Kenton Lee, Luheng He, M.~Lewis, and Luke Zettlemoyer. 2017.
\newblock End-to-end neural coreference resolution.
\newblock In \emph{EMNLP}.

\bibitem[{Lee et~al.(2018)Lee, He, and Zettlemoyer}]{Lee2018HigherorderCR}
Kenton Lee, Luheng He, and Luke Zettlemoyer. 2018.
\newblock Higher-order coreference resolution with coarse-to-fine inference.
\newblock In \emph{NAACL-HLT}.

\bibitem[{Liu and Lapata(2019)}]{liu-lapata-2019-hierarchical}
Yang Liu and Mirella Lapata. 2019.
\newblock \href {https://doi.org/10.18653/v1/P19-1500} {Hierarchical
  transformers for multi-document summarization}.
\newblock In \emph{Proceedings of the 57th Annual Meeting of the Association
  for Computational Linguistics}, pages 5070--5081, Florence, Italy.
  Association for Computational Linguistics.

\bibitem[{Luo(2005)}]{Luo2005OnCR}
Xiaoqiang Luo. 2005.
\newblock On coreference resolution performance metrics.
\newblock In \emph{HLT/EMNLP}.

\bibitem[{Ma et~al.(2020)Ma, Wang, Anubhai, Ballesteros, and
  Al-Onaizan}]{Ma2020ResourceEnhancedNM}
Jie Ma, Shuai Wang, Rishita Anubhai, Miguel Ballesteros, and Yaser Al-Onaizan.
  2020.
\newblock Resource-enhanced neural model for event argument extraction.
\newblock \emph{EMNLP}, abs/2010.03022.

\bibitem[{Martschat and Strube(2015)}]{martschat2015latent}
Sebastian Martschat and Michael Strube. 2015.
\newblock Latent structures for coreference resolution.
\newblock \emph{Transactions of the Association for Computational Linguistics},
  3:405--418.

\bibitem[{McDonald and Nivre(2007)}]{McDonald2007CharacterizingTE}
R.~McDonald and Joakim Nivre. 2007.
\newblock Characterizing the errors of data-driven dependency parsing models.
\newblock In \emph{EMNLP-CoNLL}.

\bibitem[{McDonald and Nivre(2011)}]{McDonald2011AnalyzingAI}
R.~McDonald and Joakim Nivre. 2011.
\newblock Analyzing and integrating dependency parsers.
\newblock \emph{Computational Linguistics}, 37:197--230.

\bibitem[{Meged et~al.(2020)Meged, Caciularu, Shwartz, and
  Dagan}]{Meged2020ParaphrasingVC}
Y.~Meged, Avi Caciularu, Vered Shwartz, and I.~Dagan. 2020.
\newblock Paraphrasing vs coreferring: Two sides of the same coin.
\newblock \emph{ArXiv}, abs/2004.14979.

\bibitem[{Ng(2010)}]{ng2010supervised}
Vincent Ng. 2010.
\newblock Supervised noun phrase coreference research: The first fifteen years.
\newblock In \emph{Proceedings of the 48th annual meeting of the association
  for computational linguistics}, pages 1396--1411.

\bibitem[{Pandian et~al.(2018)Pandian, Mulaffer, Oflazer, and
  AlZeyara}]{Pandian2018EventCR}
A.~Pandian, Lamana Mulaffer, K.~Oflazer, and A.~AlZeyara. 2018.
\newblock Event coreference resolution using neural network classifiers.
\newblock \emph{ArXiv}, abs/1810.04216.

\bibitem[{Pennington et~al.(2014)Pennington, Socher, and
  Manning}]{Pennington2014GloveGV}
Jeffrey Pennington, R.~Socher, and Christopher~D. Manning. 2014.
\newblock Glove: Global vectors for word representation.
\newblock In \emph{EMNLP}.

\bibitem[{Raghunathan et~al.(2010)Raghunathan, Lee, Rangarajan, Chambers,
  Surdeanu, Jurafsky, and Manning}]{raghunathan2010multi}
Karthik Raghunathan, Heeyoung Lee, Sudarshan Rangarajan, Nathanael Chambers,
  Mihai Surdeanu, Dan Jurafsky, and Christopher~D Manning. 2010.
\newblock A multi-pass sieve for coreference resolution.
\newblock In \emph{Proceedings of the 2010 Conference on Empirical Methods in
  Natural Language Processing}, pages 492--501.

\bibitem[{Surdeanu et~al.(2007)Surdeanu, i~Villodre, Carreras, and
  Comas}]{Surdeanu2007CombinationSF}
M.~Surdeanu, Llu{\'i}s~M{\`a}rquez i~Villodre, X.~Carreras, and P.~Comas. 2007.
\newblock Combination strategies for semantic role labeling.
\newblock \emph{Journal of Artificial Intelligence Research}, 29.

\bibitem[{Upadhyay et~al.(2016)Upadhyay, Gupta, Christodoulopoulos, and
  Roth}]{Upadhyay2016RevisitingTE}
Shyam Upadhyay, Nitish Gupta, Christos Christodoulopoulos, and D.~Roth. 2016.
\newblock Revisiting the evaluation for cross document event coreference.
\newblock In \emph{COLING}.

\bibitem[{Vilain et~al.(1995)Vilain, Burger, Aberdeen, Connolly, and
  Hirschman}]{Vilain1995AMC}
Marc~B. Vilain, J.~Burger, J.~Aberdeen, D.~Connolly, and L.~Hirschman. 1995.
\newblock A model-theoretic coreference scoring scheme.
\newblock In \emph{MUC}.

\bibitem[{Wang et~al.(2017)Wang, Hamza, and Florian}]{Wang2017BilateralMM}
Z.~Wang, W.~Hamza, and Radu Florian. 2017.
\newblock Bilateral multi-perspective matching for natural language sentences.
\newblock In \emph{IJCAI}.

\bibitem[{Welbl et~al.(2018)Welbl, Stenetorp, and
  Riedel}]{Welbl2018ConstructingDF}
Johannes Welbl, Pontus Stenetorp, and S.~Riedel. 2018.
\newblock Constructing datasets for multi-hop reading comprehension across
  documents.
\newblock \emph{Transactions of the Association for Computational Linguistics},
  6:287--302.

\bibitem[{Wiseman et~al.(2016)Wiseman, Rush, and Shieber}]{wiseman2016learning}
Sam Wiseman, Alexander~M Rush, and Stuart~M Shieber. 2016.
\newblock Learning global features for coreference resolution.
\newblock \emph{arXiv preprint arXiv:1604.03035}.

\bibitem[{Xu and Choi(2020)}]{Xu2020RevealingTM}
Liyan Xu and Jinho~D. Choi. 2020.
\newblock Revealing the myth of higher-order inference in coreference
  resolution.
\newblock In \emph{EMNLP}.

\bibitem[{Yan et~al.(2019)Yan, Xia, Wu, Bi, Zhao, Zhang, Si, Wang, Wang, and
  Chen}]{Yan2019ADC}
Ming Yan, Jiangnan Xia, Chen Wu, B.~Bi, Z.~Zhao, Ji~Zhang, L.~Si, Rui Wang,
  W.~Wang, and Haiqing Chen. 2019.
\newblock A deep cascade model for multi-document reading comprehension.
\newblock In \emph{AAAI}.

\bibitem[{Yang et~al.(2015)Yang, Cardie, and Frazier}]{Yang2015AHD}
B.~Yang, Claire Cardie, and P.~Frazier. 2015.
\newblock A hierarchical distance-dependent bayesian model for event
  coreference resolution.
\newblock \emph{Transactions of the Association for Computational Linguistics},
  3:517--528.

\bibitem[{Yu et~al.(2020)Yu, Yin, and Roth}]{Yu2020PairedRL}
Xiaodong Yu, Wenpeng Yin, and D.~Roth. 2020.
\newblock Paired representation learning for event and entity coreference.
\newblock \emph{ArXiv}, abs/2010.12808.

\bibitem[{Zeng et~al.(2020)Zeng, Jin, Guan, Guo, and Cheng}]{Zeng2020EventCR}
Yutao Zeng, Xiaolong Jin, Saiping Guan, J.~Guo, and Xueqi Cheng. 2020.
\newblock Event coreference resolution with their paraphrases and
  argument-aware embeddings.
\newblock In \emph{COLING}.

\bibitem[{Zhou et~al.(2020)Zhou, Pappas, and Smith}]{Zhou2020MultilevelTA}
Xuhui Zhou, Nikolaos Pappas, and Noah~A. Smith. 2020.
\newblock Multilevel text alignment with cross-document attention.
\newblock In \emph{EMNLP}.

\end{thebibliography}
\bibliographystyle{acl_natbib}

\newpage
\appendix

\section{Implementation details}
The dataset is available here: \url{http://www.newsreader-project.eu/results/data/the-ecb-corpus/}.

Our models have approximately $9$million parameters and are trained with one Tesla V100-SXM2 GPU.

We evaluate our models using three coreference metrics. MUC counts discrepencies in links between the gold and predicted clusters (and thus ignores singletons). $B^3$ computes, for each mention $m$, the difference between the gold cluster containing $m$ and the predicted cluster containing $m$. Finally, CEAF-$e$ finds the injective alignment between predicted and gold clusters that gives the highest similarity under a defined function. For more details on metrics, refer to~\citet{Cai2010EvaluationMF}.

We report validation results for both entity (Table~\ref{tab:entitycorefdev}) and event coreference (Table~\ref{tab:eventcorefdev}).

\begin{table*}[t]
    \centering
    \scalebox{0.95}{
    \begin{tabular}{l|rrr|r}
        \hline
        & MUC & $B^3$  & CEAF-$e$ & CoNLL F1\\ \hline
        BERT-Rep & 43.4 & 40.0 & 35.9 & 39.8\\
        BERT-Rep-Xdoc & 82.7 & 67.6 & 56.1 & 68.8\\ 
        SeqXdoc & 89.1 & 76.0 & 71.1 & 78.7
        \\ 
        + Adapt & 90.2 & 77.5 & 71.7 & 79.8\\
        \hline
    \end{tabular}}
    \caption{Entity coreference F1 on ECB+ dev set, combined within- and cross-document scores using predicted document clusters. }
    \label{tab:entitycorefdev}
\end{table*}
\begin{table*}[t]
    \centering
    \scalebox{0.95}{
    \begin{tabular}{l|rrr|r}
        \hline
        & MUC & $B^3$  & CEAF-$e$ & CoNLL F1\\ \hline
        BERT-Rep & 32.2 & 50.6 & 47.5 & 43.4 \\
        BERT-Rep-Xdoc & 78.4 & 74.6 & 65.0 & 72.7\\ 
        SeqXdoc & 84.8 & 79.1 & 74.4 & 79.4
        \\ 
        + Adapt & 85.8 & 81.1 & 73.7 & 80.2\\
        \hline
    \end{tabular}}
    \caption{Event coreference F1 on ECB+ dev set, combined within- and cross-document scores using predicted document clusters. }
    \label{tab:eventcorefdev}
\end{table*}

\section{Feature Ablation}
\label{sec:appendix}
We experiment with $300$-dimensional pre-trained GloVE~\citep{Pennington2014GloveGV} embeddings in our model. Following~\citep{Barhom2019RevisitingJM}, we use both GloVE and BERT in the entity mention representations (for entity coreference) and event trigger representations (for event coreference). In the argument representations we use only GloVE.

Let $x = \langle e, V\rangle$ be an entity in document $d$ and let $s_{d_1}, \hdots, s_{d_m}$ be the static GloVE embeddings for the tokens in $d$. First we apply a non-linear transformation to each $\widetilde{s_{d_i}} = \tanh (W_t s_{d_i} + b_t)$ where $W_t \in \R^{1536 \times 300}$. Then, we take the average of $\widetilde{s_{d_i}}$ to obtain $s_{\text{CLS}}$, a static document representation. Next we extract the representation for the entity $x$ as in section~\ref{sec:entityrep}, $s_x$. Finally, we combine these representations with the BERT representations
\begin{align*}
    z_x &= \tanh(W_x^s s_x + W_x^B h_x + b_x) \\
    z_{\text{CLS}} &= \tanh(W_{\text{CLS}}^s s_{\text{CLS}} + W_{\text{CLS}}^B h_x + b_{\text{CLS}})\\
    h_{c} &= W_x z_x + W_{\text{CLS}} z_{\text{CLS}} + b_c
\end{align*}
where $h_c$ is the representation (as in \S~\ref{eq:clscomp}) and $W_x^s, W_{\text{CLS}}^s, W_x^B, W_{\text{CLS}}^B \in \R^{1536 \times 1536}$ and $b_x, b_{\text{CLS}}, b_c \in \R^{1536}$ are learned parameters.

\end{document}